\documentclass[10pt,twocolumn,letterpaper]{article}

\usepackage{graphicx}
\usepackage{amsmath}
\usepackage{amssymb}
\usepackage{booktabs}
\usepackage{tabularx}
\usepackage{colortbl}
\usepackage{multirow}
\usepackage{algorithm}
\usepackage{algpseudocode}
\usepackage[most]{tcolorbox}
\usepackage{array}
\usepackage{float}
\usepackage{needspace}
\definecolor{promptblue}{HTML}{CBE0FF}
\definecolor{promptborder}{HTML}{9CB9DD}
\newtcolorbox{promptbox}[1]{
  enhanced, breakable,
  colback=promptblue!12!white, colframe=promptborder,
  colbacktitle=promptblue, coltitle=black,
  title={#1}, title after break={#1 (continued)},
  fonttitle=\bfseries\small, fontupper=\small\raggedright,
  boxrule=0.45pt, arc=2pt, titlerule=0pt,
  toptitle=4pt, bottomtitle=4pt,
  left=8pt, right=8pt, top=7pt, bottom=7pt,
  before skip=6pt, after skip=8pt,
  before upper={\setlength{\parindent}{0pt}\setlength{\parskip}{4pt}}
}
\lstdefinestyle{skilljson}{
  basicstyle=\ttfamily\footnotesize,
  breaklines=true, breakatwhitespace=true,
  columns=fullflexible, keepspaces=true,
  showstringspaces=false, aboveskip=0pt, belowskip=0pt
}

\usepackage[pagenumbers]{cvpr}
\usepackage[T1]{fontenc}
\usepackage{lmodern}

\usepackage{cuted}
\usepackage{currfile}
\usepackage{caption}

\newcommand{\lz}[1]{#1}

\definecolor{cvprblue}{rgb}{0.21,0.49,0.74}
\definecolor{remvpblue}{HTML}{D5E6FF}
\usepackage[pagebackref,breaklinks,colorlinks,allcolors=cvprblue]{hyperref}

\def\confName{CVPR}
\def\confYear{2027}

\title{How to Learn from Videos through Recursive Self-Improvement for Socially Proactive Intelligence?}

\author{
Xinyi Che\quad Kuofei Fang\quad Zheng Lian\thanks{Corresponding author.}\\
{\small State Key Laboratory of Autonomous Intelligent Unmanned Systems}\\
{\small Tongji University}
}

\hypersetup{
  pdftitle={How to Learn from Videos through Recursive Self-Improvement for Socially Proactive Intelligence?},
  pdfauthor={Xinyi Che, Kuofei Fang, Zheng Lian}
}
\begin{document}
\maketitle

\begin{abstract}

Social proactive intelligence (SPI) requires models to understand visual contexts and infer socially appropriate next actions.
In video-grounded settings,SPI requires reasoning not only about visual cues, such as human activities, objects, and events, but also about implicit social factors, including interpersonal relationships and social norms.
Prior studies have explored extracting experiences from historical examples and retrieving relevant ones to enhance SPI judgments on subsequent tasks.
However, such approaches rely on maintaining experience repositories and accurately retrieving experiences relevant to the current social context.
To address these limitations, we propose \textbf{ReMVP}, which generates video-specific skills under the guidance of a skill generation policy.
Specifically, ReMVP employs a recursive self-improvement framework that dynamically updates the skill generation policy using feedback from trajectories.
Extensive experiments demonstrate that ReMVP enhances the ability of models to  understand diverse contextual factors in videos and improve task performance.

\end{abstract}

\section{Introduction}
\label{sec:intro}

Embodied agents increasingly connect visual observations with actions in physical environments~\cite{driess2023palme,zitkovich2023rt2}. Beyond following explicit instructions, socially proactive agents must infer whether, when, and how to assist people from the surrounding context~\cite{fang2026roboteq}. We refer to this line of research as \textbf{Social Proactive Intelligence (SPI)}. Such decisions require understanding both visible events and social expectations. Approaching someone may offer useful assistance, but may also interrupt an ongoing activity or intrude on personal space. SPI provides embodied agents with the foundation to act and think like humans in social environments.

\begin{figure}[t]
    \centering
    \includegraphics[width=\columnwidth]{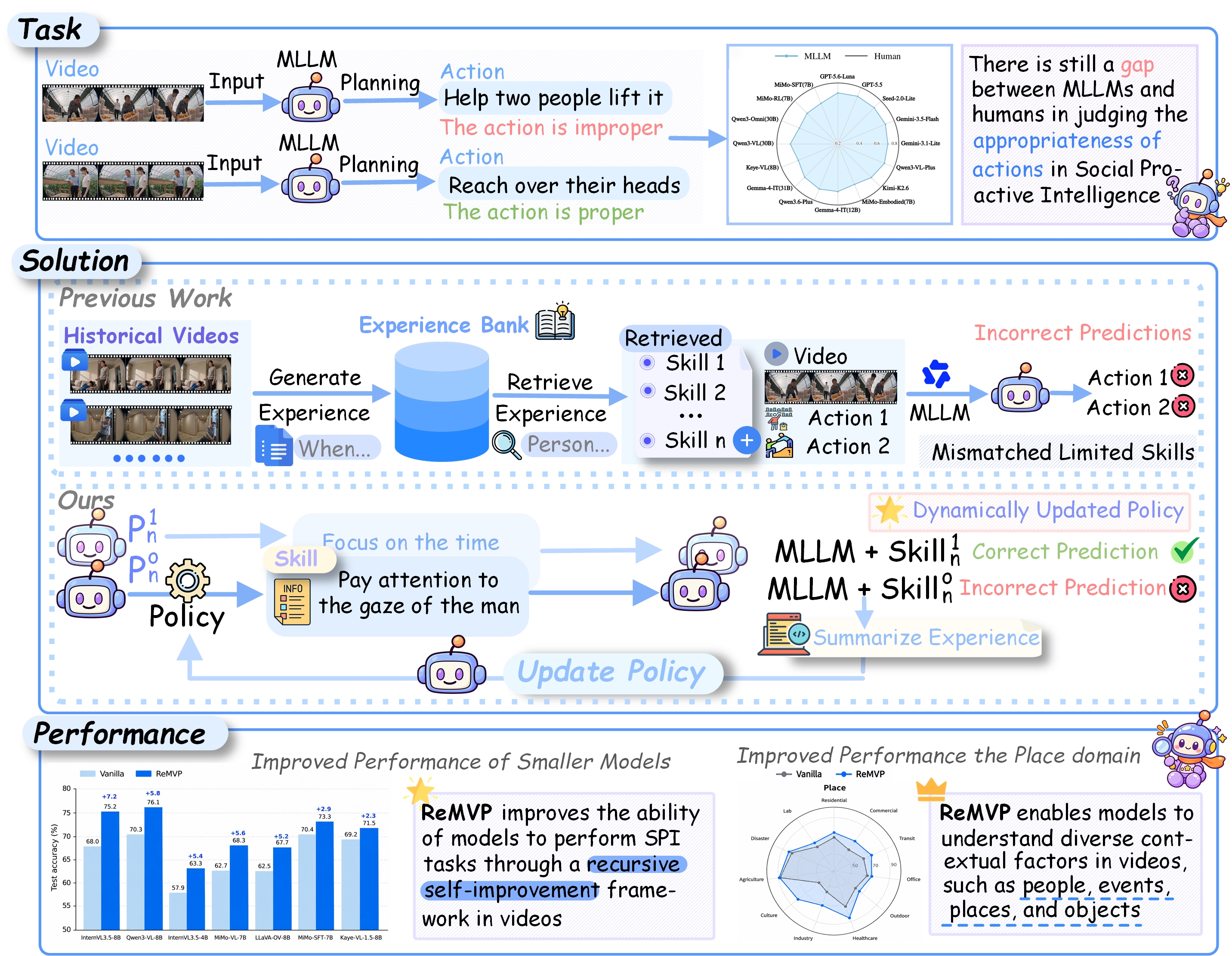}

    \caption{\lz{\textbf{Overview of ReMVP.} The top panel illustrates the SPI task definition, highlighting the performance gap between MLLMs and humans. The middle panel contrasts prior methods that generate and retrieve experiences with our recursive policy improvement framework for generating video-specific skills. The bottom panel summarizes the resulting gains in both SPI benchmark performance and video-context understanding.}}
    \label{fig:overview}
\end{figure}

However, current research indicates that embodied agents do not inherently possess SPI capability~\cite{fang2026roboteq}. They struggle to capture the visual and social context required for decision making. Existing SPI research addresses this by constructing experience libraries and retrieving from them to enhance model understanding~\cite{fang2026roboteq,rezaei2025egonormia}. Retrieval-augmented methods condition outputs on externally stored knowledge, filling factual and procedural gaps that models cannot infer from raw inputs alone~\cite{lewis2020rag,asai2024selfrag}. Agent-based approaches extend this idea by extracting experiences from historical task trajectories and transforming them into reusable guidance for new scenarios~\cite{park2023generativeagents,zhao2024expel,wang2024voyager,wang2025awm,liu2025cer}. These methods enable agents to accumulate social knowledge over time rather than relying solely on pretrained priors.
However, retrieval-augmented approaches face growing challenges as SPI tasks grow in complexity. Social scenarios are inherently diverse and context-dependent, with social norms shifting across different people, activities, and environments. Skills generated for one scenario may be mismatched when transferred to another, offering limited guidance or even conflicting signals for the current situation. Moreover, scaling up this paradigm requires retrieving from and maintaining a large experience library. As the library expands, identifying entries that truly align with the nuanced visual and social cues of a new video becomes increasingly difficult, and retrieval overhead continues to grow without guaranteed relevance.

To address this concern, we propose \textbf{ReMVP}, a framework that strengthens embodied agents' understanding of SPI and social behavior reasoning by generating scene-specific textual cues via a generative policy. ReMVP continuously refines its skill generation strategy through \textbf{Recursive Self-Improvement }(RSI), enabling the framework to evolve iteratively. Specifically, for each embodied scenario, ReMVP generates a textual prompt termed a skill, which captures relevant social norms, human intent, and contextual triggers. This skill is injected into the agent's context during inference to enhance judgment of socially appropriate actions. To guarantee the correctness and applicability of the generation policy, ReMVP summarizes and evaluates inference trajectories, using the RSI framework to update itself and emulate the progressive improvement of human understanding of social behavior theories. To the best of our knowledge, ReMVP is the first framework to introduce recursive self-improvement into SPI by dynamically refining skill generation policies.

We evaluate ReMVP extensively on the SPI benchmark. Across multiple open-source vision-language backbones, our framework consistently achieves state-of-the-art performance. Beyond main comparisons, we conduct comprehensive generalization experiments that demonstrate the effectiveness of ReMVP and the complementary interplay between its generation and retrieval components. Our work makes the following contributions:
\lz{
\begin{itemize}
    \item  We propose ReMVP, a recursive self-improvement framework that dynamically updates the skill generation policy using feedback from trajectories, enabling models to better understand complex video contexts.
    \item  We introduce a policy-guided skill generation mechanism that enables models to generate skills from training examples and leverage them to enhance their own reasoning.
    \item  Extensive experiments demonstrate that ReMVP consistently outperforms baselines across multiple backbones.
\end{itemize}
}

\begin{figure*}[t]
    \centering
    \includegraphics[width=\textwidth,trim=462bp 217bp 52bp 368bp,clip]{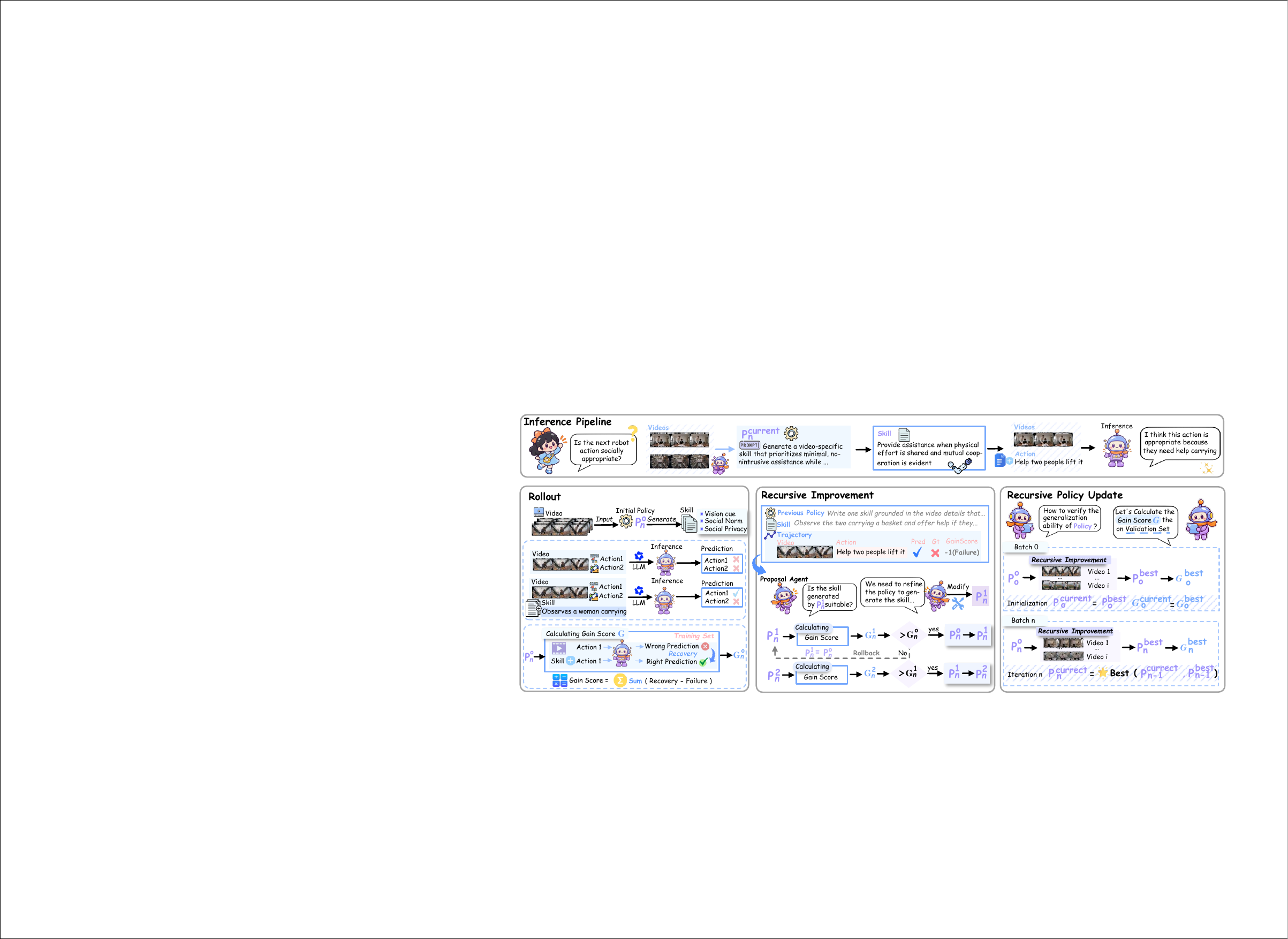}

    \caption{\lz{\textbf{Overview of ReMVP.} The top panel illustrates the inference pipeline, where the evolved policy guides video-specific skill generation for action judgment. The bottom panels present three components. \textbf{Rollout} collects paired action judgments with and without skills and computes the Gain Score. \textbf{Recursive Improvement} iteratively refines the skill generation policy based on trajectory feedback. \textbf{Recursive Policy Update} selects and retains the best policy through validation across training batches.}}
    \label{fig:method_overview}
\end{figure*}

\section{Related Work}
\label{sec:related}

\subsection{Social Proactive Intelligence}
Early research on social intelligence investigates everyday norms and the social appropriateness of robot behavior, mostly from static scenarios and population-level expectations~\cite{forbes2020socialchemistry,tjomsland2022manners}. RobotEQ formalizes this line of work as Social Proactive Intelligence (SPI), pushing embodied agents from reactive task execution toward understanding human needs, social norms, and contextual constraints in diverse scenes~\cite{fang2026roboteq}. RobotEQ-Video extends SPI evaluation from images to dynamic videos with a fine-grained world-state taxonomy~\cite{che2026roboteqvideo}, while RobotEQ 3.0 further incorporates individual user profiles for personalized SPI~\cite{3.0}.
Beyond evaluation, prior methods improve socially appropriate decision-making by retrieving external social knowledge or historical examples~\cite{rezaei2025egonormia}. In contrast, ReMVP recursively evolves a skill generation policy using feedback from labeled action judgments, producing video-specific guidance without relying on retrieved experiences.

\subsection{Recursive Self-Improvement}
Feedback-driven self-improvement enables language agents to refine their behavior without updating model parameters. Existing approaches leverage self-generated feedback, verbal reflections, and accumulated experiences to improve subsequent decisions~\cite{madaan2023selfrefine,shinn2023reflexion,zhao2024expel,wang2024voyager}. Beyond refining individual outputs, prompt optimization directly improves the instructions governing model behavior. Methods such as PromptAgent~\cite{wang2024promptagent}, Promptbreeder~\cite{fernando2024promptbreeder}, and GEPA~\cite{agrawal2026gepa} iteratively optimize prompts through error feedback, evolutionary search, and trajectory reflection.

Building on these advances, ReMVP introduces recursive policy refinement for video-grounded SPI. Unlike methods that directly optimize task prompts or retrieve past experiences, ReMVP evolves a reusable skill generation policy that produces video-specific guidance. Each policy revision is evaluated through paired action judgments with and without generated skills, while validation across batches guides the retention of beneficial revisions.

\section{Method}
\label{sec:method}

To improve social reasoning and action judgment in videos, we propose ReMVP, an agent framework based on recursive self-improvement (RSI) that uses feedback from rollout trajectories to iteratively update a policy for skill generation. This policy guides the generation of video-specific skills, helping the model interpret visual evidence and reason about socially appropriate actions. As illustrated in Figure~\ref{fig:method_overview}, ReMVP comprises two components: \emph{Inference Pipeline} and \emph{RSI Process}.

\subsection{Task Definition}
\label{sec:formulation}

RobotEQ-Video~\cite{che2026roboteqvideo} is a full-modality dataset designed to evaluate SPI, specifically testing models' ability to judge the social appropriateness of robot actions based on video observations and textual action candidates.

For each video $v_i$, we are given an ordered list of $m_i$ candidate robot actions $\mathcal{A}_i = (a_{i1}, \ldots, a_{im_i})$, and a fixed judgment question $Q$. The model $f_\theta$ takes these inputs and produces a vector of binary predictions:

\begin{equation}
    \hat{\mathbf{y}}_i = f_\theta(v_i, \mathcal{A}_i, Q, s_i).
    \label{eq:task_judgment}
\end{equation}

Here, $\hat{\mathbf{y}}_i = (\hat{y}_{i1}, \ldots, \hat{y}_{im_i}) \in \{0,1\}^{m_i}$, where each $\hat{y}_{ij}$ indicates whether action $a_{ij}$ is socially appropriate in the video. The dataset provides gold labels $\mathbf{y}_i = (y_{i1}, \ldots, y_{im_i})$ for feedback and evaluation.

\subsection{Inference Pipeline}
\label{sec:inference}

Humans rarely judge social behavior from scratch. They first recall relevant norms, interpret the scene, and form a mental sketch of appropriate conduct before acting. We mimic this cognitive habit by introducing a skill generation step before action judgment. We apply an evolved policy $p_n^{\mathrm{current}}$, retained after $n$ rounds of recursive self-improvement as described in Section~\ref{sec:best_policy_update}, to produce a video-specific skill for each input. This skill compiles clues and contextual evidence relevant to SPI, such as social norms, human intent, and interpersonal boundaries, serving as a proxy for human-like understanding of the scene and its social dynamics.

Specifically, the policy guides the model to generate a video-grounded skill based on the input video,

\begin{equation}
    s_i = G_{\theta}(v_i; p_n^{\mathrm{current}}).
    \label{eq:inference_skill}
\end{equation}

Here, $s_i$ provides decision guidance adapted to the current video context, while $p_n^{\mathrm{current}}$ captures general reasoning strategies refined through RSI. The skill output format is provided in Appendix A.2. The model then conditions its action judgments on the generated skill:

\begin{equation}
    \hat{\mathbf{y}}_i = f_{\theta}(v_i, \mathcal{A}_i, Q, s_i).
    \label{eq:inference_judgment}
\end{equation}

\subsection{Recursive Self-Improvement}

Initial policies designed by humans may be incomplete or suboptimal. To address this limitation, we introduce recursive self-improvement (RSI), enabling agents to iteratively refine their skill generation policies using feedback from past trajectories and progressively improve their reasoning about complex social contexts.

\subsubsection{Rollout}
\label{sec:policy_proposal}

We first sample past inference trajectories to construct training samples that drive policy evolution. At the start of each update $n \geq 1$, we set the initial policy for the current round $p_n^0$ to the best strategy retained from all previous rounds, denoted $p_{n-1}^{\mathrm{current}}$. This policy captures the optimal skill generation instruction evolved up to update $n-1$, giving

\begin{equation}
    p_n^{0} = p_{n-1}^{\mathrm{current}}.
    \label{eq:update_policy_init}
\end{equation}

For the very first update at $n=0$, no evolved strategy is available. We instead initialize with a human-crafted instruction $p_{\mathrm{init}}$, so $p_0^0 = p_{\mathrm{init}}$. The detailed content of this initial prompt is provided in Appendix A.1.
For each video $v_i$ in update $n$, we first perform a rollout without skill guidance. Given the video, the complete candidate action list $\mathcal{A}_i$, and the judgment question $Q$, the model composes these inputs into a response trajectory through $f_\theta$,

\begin{equation}
    R_i \triangleq \bigl(f_\theta(v_i,\, \mathcal{A}_i,\, Q),\, v_i,\, \mathcal{A}_i,\, Q\bigr).
    \label{eq:bare_rollout}
\end{equation}

This trajectory is not merely a flat prediction but a composite output jointly determined by the video content, action candidates, and the question formulation.

We then perform a rollout guided by the generated skill. Specifically, the initial policy for the current update, $p_n^0$, guides the model to generate a video-specific skill,

\begin{equation}
    s_i = G_{\theta}(v_i; p_n^0).
    \label{eq:rollout_skill}
\end{equation}
To assess the effect of the generated skill on action judgments, we perform a second rollout using the same video, candidate actions, and judgment question,

\begin{equation}
    R_i^s \triangleq \bigl(f_\theta(v_i,\, \mathcal{A}_i,\, Q,\, s_i),\, v_i,\, \mathcal{A}_i,\, Q,\, s_i\bigr).
    \label{eq:skill_rollout}
\end{equation}

Here, $R_i$ and $R_i^s$ denote the response trajectories without and with skill guidance, respectively. Both trajectories contain judgments for all candidate actions in their original order. Skill generation and action judgment share the same frozen model parameters $\theta$.

To evaluate the impact of generated skills on action judgments, we define a Gain Score for each training batch $D_n$. Let $\hat{y}_{ij}$ and $\hat{y}_{ij}^s$ denote predictions without and with skills, respectively, and $y_{ij}$ the ground-truth label. The corresponding correctness indicators are defined as:
\begin{equation}
    c_{ij} = \mathbf{1}\{\hat{y}_{ij} = y_{ij}\},
    \qquad
    c_{ij}^s = \mathbf{1}\{\hat{y}_{ij}^s = y_{ij}\}.
\end{equation}

Based on these correctness indicators, we compute the batch-level Recovery, Failure, and Gain Score as follows:

\begin{equation}
\begin{aligned}
    \operatorname{Recover}_n
    &= \sum_{i\in D_n}\sum_{j=1}^{m_i}
       (1-c_{ij})c_{ij}^s,\\
    \operatorname{Failure}_n
    &= \sum_{i\in D_n}\sum_{j=1}^{m_i}
       c_{ij}(1-c_{ij}^s),\\
    \operatorname{Gain}_n
    &= \operatorname{Recover}_n - \operatorname{Failure}_n.
\end{aligned}
\label{eq:gain_score}
\end{equation}

Recover counts the judgments corrected by the generated skill, whereas Failure counts those that become incorrect. The Gain Score captures the net change in correct judgments across the batch, with positive and negative values indicating overall improvement and degradation, respectively. This metric characterizes the extent to which the generated skill compensates for the model's social proactive intelligence capability.
To guide subsequent revisions of the skill generation policy, we provide this score to the proposal agent as quantitative feedback.
The details of the rollout information is shown in Appendix A.3.
\begin{table*}[t]
    \centering
    \captionsetup{font=normalsize,labelsep=colon,width=\textwidth}

    \caption{\lz{\textbf{Main Results on RobotEQ-Video.} We compare three settings across vision-language models. Vanilla  predicts action appropriateness without skills. Fixed Policy generates skills using a fixed initial policy, while ReMVP recursively refines the skill generation policy through trajectory feedback. We report Accuracy (Acc), Macro-F1 (Mac), and their average (Avg) on the synthesized subset, real subset, and full test set. All model parameters remain frozen throughout the experiments.}}
    \label{tab:main_results}
    \fontsize{7.3}{8.4}\selectfont
    \setlength{\tabcolsep}{2pt}
    \newcommand{\resulthead}[1]{{\bfseries #1}}
    \begin{tabularx}{\textwidth}{@{}p{1.50in}p{0.68in}*{9}{>{\centering\arraybackslash}X}@{}}
        \specialrule{0.4pt}{0pt}{0.8pt}
        \specialrule{0.4pt}{0pt}{1.5pt}
        & & \multicolumn{3}{c}{\resulthead{Synthesized Subset}} & \multicolumn{3}{c}{\resulthead{Real Subset}} & \multicolumn{3}{c}{\resulthead{Full Dataset}} \\
        \cmidrule(lr){3-5}\cmidrule(lr){6-8}\cmidrule(lr){9-11}
        \resulthead{Model} & \resulthead{Setting} & \resulthead{Acc} & \resulthead{Mac} & \resulthead{Avg} & \resulthead{Acc} & \resulthead{Mac} & \resulthead{Avg} & \resulthead{Acc} & \resulthead{Mac} & \resulthead{Avg} \\
        \midrule[0.4pt]
        InternVL3.5-4B-Instruct~\cite{wang2025internvl35} & Vanilla & 54.62 & 53.72 & 54.17 & 64.74 & 63.44 & 64.09 & 57.86 & 57.81 & 57.84 \\
         & Fixed Policy & 57.61 & 58.57 & 58.09 & 63.11 & 55.73 & 59.42 & 59.37 & 58.81 & 59.09 \\
        \rowcolor{remvpblue}
         & \textbf{ReMVP} & 58.97 & 58.43 & 58.70 & 73.78 & 68.41 & 71.09 & 63.73 & 62.05 & 62.89 \\
        \specialrule{0.3pt}{0pt}{1pt}
        MiMo-VL-7B-RL~\cite{xiaomi2025mimovl} & Vanilla & 60.26 & 66.56 & 63.41 & 69.44 & 70.09 & 69.76 & 63.20 & 67.74 & 65.47 \\
         & Fixed Policy & 62.99 & 68.55 & 65.77 & 67.63 & 66.77 & 67.20 & 64.48 & 68.02 & 66.25 \\
        \rowcolor{remvpblue}
         & \textbf{ReMVP} & 66.84 & 72.64 & 69.74 & 72.88 & 77.55 & 75.21 & 68.78 & 74.24 & 71.51 \\
        \specialrule{0.3pt}{0pt}{1pt}
        MiMo-VL-7B-SFT~\cite{xiaomi2025mimovl} & Vanilla & 71.37 & 72.40 & 71.88 & 74.50 & 72.33 & 73.42 & 72.37 & 72.41 & 72.39 \\
         & Fixed Policy & 70.26 & 69.16 & 69.71 & 70.71 & 68.11 & 69.41 & 70.40 & 68.88 & 69.64 \\
        \rowcolor{remvpblue}
         & \textbf{ReMVP} & 72.82 & 74.72 & 73.77 & 74.32 & 75.78 & 75.05 & 73.30 & 75.09 & 74.20 \\
        \specialrule{0.3pt}{0pt}{1pt}
        VideoLLaMA3-7B~\cite{zhang2025videollama3} & Vanilla & 56.41 & 59.86 & 58.13 & 64.20 & 64.70 & 64.45 & 58.91 & 61.73 & 60.32 \\
         & Fixed Policy & 60.34 & 65.84 & 63.09 & 54.79 & 63.00 & 58.89 & 58.56 & 65.02 & 61.79 \\
        \rowcolor{remvpblue}
         & \textbf{ReMVP} & 63.85 & 64.83 & 64.34 & 64.56 & 66.07 & 65.31 & 64.07 & 65.25 & 64.66 \\
        \specialrule{0.3pt}{0pt}{1pt}
        Qwen3-VL-8B-Instruct~\cite{bai2025qwen3vl} & Vanilla & 68.89 & 68.34 & 68.61 & 73.24 & 70.96 & 72.10 & 70.28 & 69.31 & 69.80 \\
         & Fixed Policy & 69.15 & 68.55 & 68.85 & 75.95 & 72.13 & 74.04 & 71.33 & 69.99 & 70.66 \\
        \rowcolor{remvpblue}
         & \textbf{ReMVP} & 74.96 & 72.57 & 73.76 & 78.66 & 75.49 & 77.08 & 76.15 & 73.51 & 74.83 \\
        \specialrule{0.3pt}{0pt}{1pt}
        Keye-VL-1.5-8B~\cite{yang2025keyevl15} & Vanilla & 69.66 & 69.27 & 69.47 & 68.17 & 67.16 & 67.67 & 69.18 & 68.63 & 68.91 \\
         & Fixed Policy & 69.06 & 68.66 & 68.86 & 66.91 & 64.43 & 65.67 & 68.37 & 67.49 & 67.93 \\
        \rowcolor{remvpblue}
         & \textbf{ReMVP} & 70.60 & 69.23 & 69.91 & 74.50 & 71.70 & 73.10 & 71.85 & 70.05 & 70.95 \\
        \specialrule{0.3pt}{0pt}{1pt}
        InternVL3.5-8B-Instruct~\cite{wang2025internvl35} & Vanilla & 66.41 & 63.49 & 64.95 & 71.25 & 66.07 & 68.66 & 67.96 & 64.37 & 66.17 \\
         & Fixed Policy & 69.66 & 69.03 & 69.34 & 72.88 & 70.87 & 71.87 & 70.69 & 69.70 & 70.19 \\
        \rowcolor{remvpblue}
         & \textbf{ReMVP} & 73.59 & 71.87 & 72.73 & 78.12 & 74.27 & 76.19 & 75.04 & 72.72 & 73.88 \\
        \specialrule{0.3pt}{0pt}{1pt}
        LLaVA-OneVision-2-8B-Instruct~\cite{an2026llavaonevision2} & Vanilla & 61.37 & 45.02 & 53.19 & 64.92 & 45.33 & 55.13 & 62.51 & 45.16 & 53.84 \\
         & Fixed Policy & 60.85 & 44.58 & 52.72 & 65.28 & 47.79 & 56.54 & 62.28 & 45.59 & 53.93 \\
        \rowcolor{remvpblue}
         & \textbf{ReMVP} & 67.52 & 58.48 & 63.00 & 67.99 & 56.49 & 62.24 & 67.67 & 57.96 & 62.82 \\
        \specialrule{0.4pt}{0pt}{0.8pt}
        \specialrule{0.4pt}{0pt}{0pt}
    \end{tabularx}

\end{table*}

\subsubsection{Recursive Improvement}
\label{sec:recursive_improvement}

At update $n$, the initial policy $p_n^0$ produces skills and rollouts on batch $D_n$. To refine this policy, we use the frozen model itself as a proposal agent $\mathcal{A}_\theta^P$. The agent takes the current policy, the generated skills, the corresponding rollouts, and the Gain Score as input to identify weaknesses and propose revisions:

\begin{equation}
    p_n^{1} = \mathcal{A}_\theta^P\bigl(p_n^0,\, \mathbf{S}_n^0,\, \mathbf{R}_n,\, \mathbf{R}_n^s,\, \operatorname{Gain}_n^0\bigr).
    \label{eq:policy_proposal}
\end{equation}

Here, $\mathbf{S}_n^0$ collects the skills generated for batch $D_n$ under $p_n^0$, while $\mathbf{R}_n$ and $\mathbf{R}_n^s$ collect the rollouts without and with skill guidance, respectively. The boldface notation indicates that these are vectors aggregating all data within the batch. $\operatorname{Gain}_n^0$ is the Gain Score computed from these trajectories.

We then evaluate the revised policy $p_n^1$ by regenerating skills and computing its Gain Score $\operatorname{Gain}_n^1$ on the same batch. If $\operatorname{Gain}_n^1 \leq \operatorname{Gain}_n^0$, the revision is discarded. If the new policy improves the score, we proceed to further evolutions, allowing up to $\mathit{extra}$ additional proposal rounds.

At each round $i = 1, \ldots, \mathit{extra}$, the proposal agent takes the current policy $p_n^i$, its skills $\mathbf{S}_n^i$, rollouts $\mathbf{R}_n^i$ and $\mathbf{R}_n^{s,i}$, and score $\operatorname{Gain}_n^i$ to produce the next candidate

\begin{equation}
    p_n^{i+1} = \mathcal{A}_\theta^P\bigl(p_n^i,\, \mathbf{S}_n^i,\, \mathbf{R}_n^i,\, \mathbf{R}_n^{s,i},\, \operatorname{Gain}_n^i\bigr).
    \label{eq:recursive_proposal}
\end{equation}

The revision is retained only if $\operatorname{Gain}_n^{i+1} > \operatorname{Gain}_n^i$. This loop continues until $i = \mathit{extra}$ or the Gain Score stops improving.
Finally, we select the policy with the highest Gain Score across all evaluated rounds as the batch-best:

\begin{equation}
    p_n^{\mathrm{best}} = p_n^{i^\star},
    \qquad
    i^\star = \underset{i \in \{0, \ldots, \mathit{extra}\}}{\arg\max}\, \operatorname{Gain}_n^i.
    \label{eq:batch_best_policy}
\end{equation}

In case of ties, the earlier policy is retained.

\subsubsection{Recursive Policy Update}
\label{sec:best_policy_update}

Multimodal video inputs impose substantial context demands, restricting the number of training examples processed in each batch. Consequently, policy improvements on a single batch may not generalize to other scenarios. To mitigate this issue, we introduce a recursive policy update mechanism that evaluates candidate policies on a fixed validation set $D_v$, ensuring that policy updates generalize beyond the training examples used for proposal generation.

Starting from the initial policy $p_0^{0}$, recursive improvement produces the best policy for the first batch, $p_0^{\mathrm{best}}$. We initialize the retained policy as:
\begin{equation}
    p_0^{\mathrm{current}}=p_0^{\mathrm{best}}.
    \label{eq:current_policy_init}
\end{equation}
For each subsequent update $n$, we initialize the within-batch policy with $p_n^{0}=p_{n-1}^{\mathrm{current}}$ and obtain $p_n^{\mathrm{best}}$ through recursive improvement. The retained policy is then updated recursively:
\begin{equation}
    p_n^{\mathrm{current}}
    =
    \operatorname{Best}_{D_v}
    \left(
        p_{n-1}^{\mathrm{current}},
        p_n^{\mathrm{best}}
    \right).
    \label{eq:recursive_policy_update}
\end{equation}
Here, $\operatorname{Best}_{D_v}$ selects the policy with the higher Gain Score on the fixed validation set $D_v$, retaining the previous policy in the event of a tie. The validation set $D_v$ is disjoint from the training batches, and its Gain Score is computed over all action judgments.

This recursive update mechanism integrates beneficial policy revisions across diverse training batches while validating their effectiveness on unseen examples, thereby mitigating overfitting to individual batches.

\section{Experiments}
\label{sec:experiments}

\subsection{Experimental Dataset}
\label{sec:exp_setup}

RobotEQ-Video~\cite{che2026roboteqvideo} is a video-centric benchmark for evaluating Social Proactive Intelligence from world-state observations. Extending prior image-based evaluation to videos, it introduces a hierarchical world-state taxonomy to capture diverse contextual factors relevant to socially appropriate robot actions. The dataset comprises synthesized and real videos. Our experiments involve 2,146 videos and 17,544 candidate actions, partitioned by video into disjoint training, validation, and test sets. Detailed statistics for synthesized and real videos are presented in Table~\ref{tab:dataset_statistics}.

\subsection{Implementation Details}
\label{sec:implementation_details}

\paragraph{Settings.}
To evaluate the effectiveness of ReMVP, we conduct experiments on RobotEQ-Video using eight open-source vision-language models ranging from 4B to 8B parameters, as listed in Table~\ref{tab:main_results}. By default, each RSI update processes a batch of four videos with $extra=1$ in main experiment, allowing one policy revision proposal per update. For each video, we uniformly sample eight frames and jointly evaluate all candidate actions. Model parameters remain frozen throughout RSI and evaluation.
We consider three experimental settings. In the \emph{Vanilla} setting, the model directly predicts action appropriateness without skills. The \emph{Fixed Policy} setting generates video-specific skills using a fixed initial policy. In contrast, \emph{ReMVP} iteratively updates skill generation policy through RSI and uses the retained policy to guide skill generation and action judgment. For update validation, we evaluate on validation set every four batches. More experiment details are provided in Appendix.

\begin{table}[b]
    \centering

    \caption{\lz{\textbf{Statistics of the experimental dataset.} The dataset is partitioned into training, validation, and test sets. For each split, we report the number of videos and candidate actions, further categorized by synthesized (Syn.) and real (Real) subsets.}}
    \label{tab:dataset_statistics}
    \fontsize{7.5}{9}\selectfont
    \setlength{\tabcolsep}{1.8pt}
    \renewcommand{\arraystretch}{0.95}
    \begin{tabular*}{\columnwidth}{@{\extracolsep{\fill}}*{9}{c}@{}}
        \specialrule{0.4pt}{0pt}{1pt}
        \specialrule{0.4pt}{0pt}{2pt}
        & \multicolumn{2}{c}{\textbf{Train}}
        & \multicolumn{2}{c}{\textbf{Val}}
        & \multicolumn{2}{c}{\textbf{Test}}
        & \multicolumn{2}{c}{\textbf{Total}} \\
        \cmidrule(lr){2-3}
        \cmidrule(lr){4-5}
        \cmidrule(lr){6-7}
        \cmidrule(lr){8-9}
        & Syn. & Real. & Syn. & Real. & Syn. & Real. & Syn. & Real. \\
        \midrule[0.4pt]
        Video  & 1044 & 789  & 53  & 47  & 121  & 92  & 1218  & 928  \\
        Action & 9630 & 5374 & 492 & 325 & 1103 & 620 & 11225 & 6319 \\
        \specialrule{0.4pt}{2pt}{1pt}
        \specialrule{0.4pt}{0pt}{0pt}
    \end{tabular*}
\end{table}

\paragraph{Evaluation Metrics.}
We report accuracy (Acc), macro-F1 (Mac), and their arithmetic mean (Avg), with Acc and Mac defined as
\begin{equation}
    \mathrm{Acc}=\frac{TP+TN}{N},\qquad
    \mathrm{Mac}=\frac{\mathrm{F1}_0+\mathrm{F1}_1}{2}.
    \label{eq:evaluation_metrics}
\end{equation}
Here, $N$ denotes the total number of labeled candidate actions, while $TP$ and $TN$ denote correctly predicted appropriate and inappropriate actions, respectively. We adopt macro-F1 to give equal weight to both action classes, providing a more balanced evaluation under class imbalance. Avg is the arithmetic mean of Acc and Mac, jointly reflecting overall accuracy and balanced recognition performance.  All metrics are reported as percentages on the synthesized, real, and full test sets. Full test set metrics are computed from pooled predictions rather than averaged subset scores.

\section{Results and Discussion}
\subsection{Main Results}

We investigate the effectiveness of ReMVP and the impact of recursive policy updates by comparing ReMVP with Vanilla and Fixed Policy across eight vision-language backbones, as shown in Table~\ref{tab:main_results}.

On the full test set, ReMVP consistently outperforms Vanilla in terms of Avg, with the largest improvement observed on LLaVA-OneVision-2-8B-Instruct.
Fixed Policy improves Avg over Vanilla on six of the eight backbones, demonstrating the potential of video-specific skills to support socially appropriate action judgments. However, performance declines on MiMo-VL-7B-SFT and Keye-VL-1.5-8B, indicating that skills generated under a fixed policy are not consistently beneficial. This observation motivates the need to adapt the skill generation policy through recursive self-improvement (RSI).
ReMVP further outperforms Fixed Policy across all eight backbones, highlighting the effectiveness of feedback-guided policy updates. By leveraging video evidence and feedback from both corrected and degraded judgments, RSI progressively improves the policy for extracting diverse contextual factors from videos without updating the underlying model parameters.
ReMVP also consistently improves Avg over Vanilla on both the synthesized and real subsets across all backbones, demonstrating its effectiveness across different data sources. Furthermore, ReMVP improves Accuracy and Macro-F1 on all eight backbones, indicating more accurate action judgments and consistent improvements.

\subsection{Effect of RSI Evolution Rounds}
\label{sec:exp_policy_evolution}

\begin{figure}[H]
    \centering
    \begin{minipage}[b]{0.49\columnwidth}
        \centering
        \includegraphics[width=\linewidth]{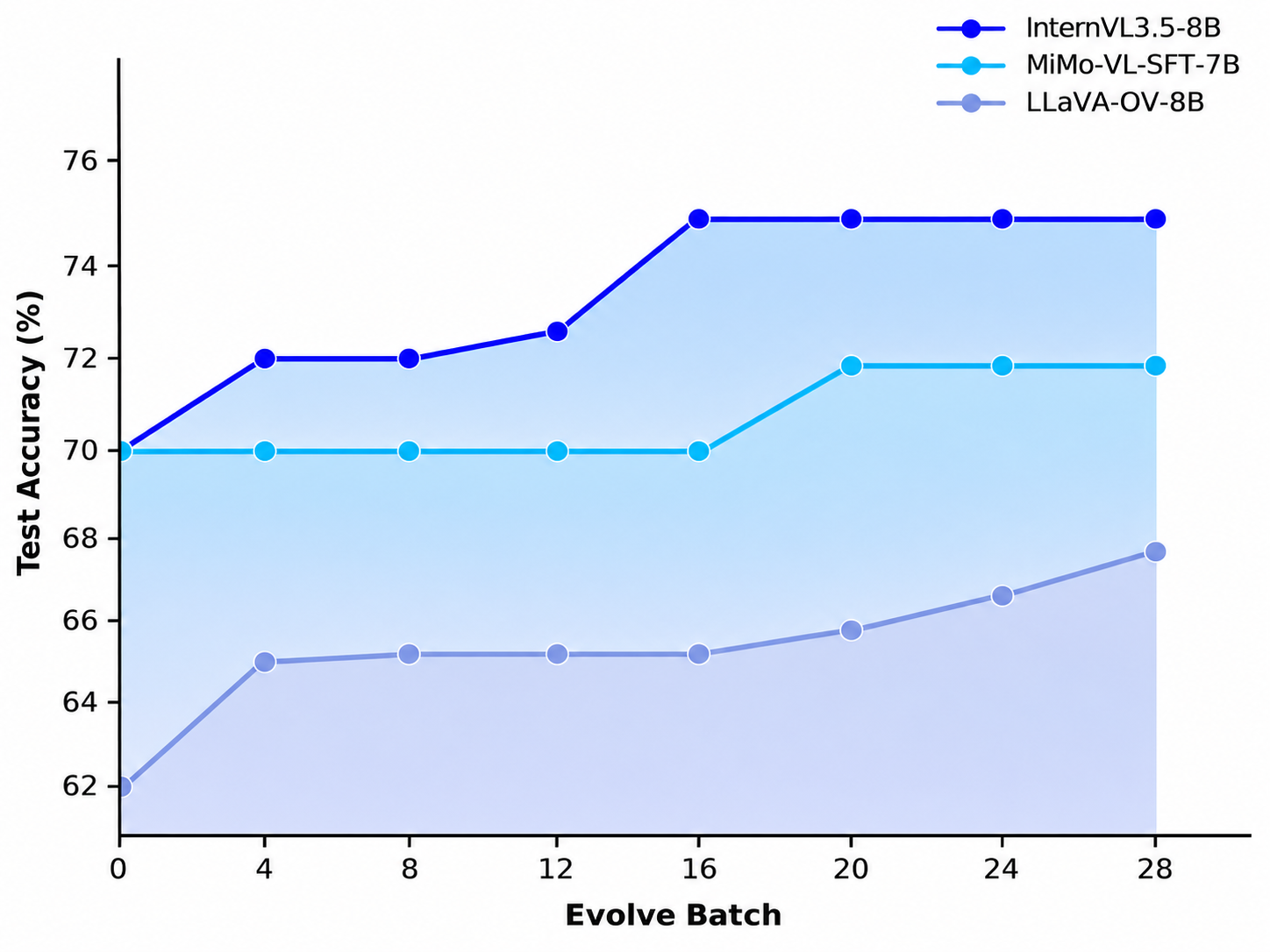}
        \par\small (a) Accuracy
    \end{minipage}\hfill
    \begin{minipage}[b]{0.49\columnwidth}
        \centering
        \includegraphics[width=\linewidth]{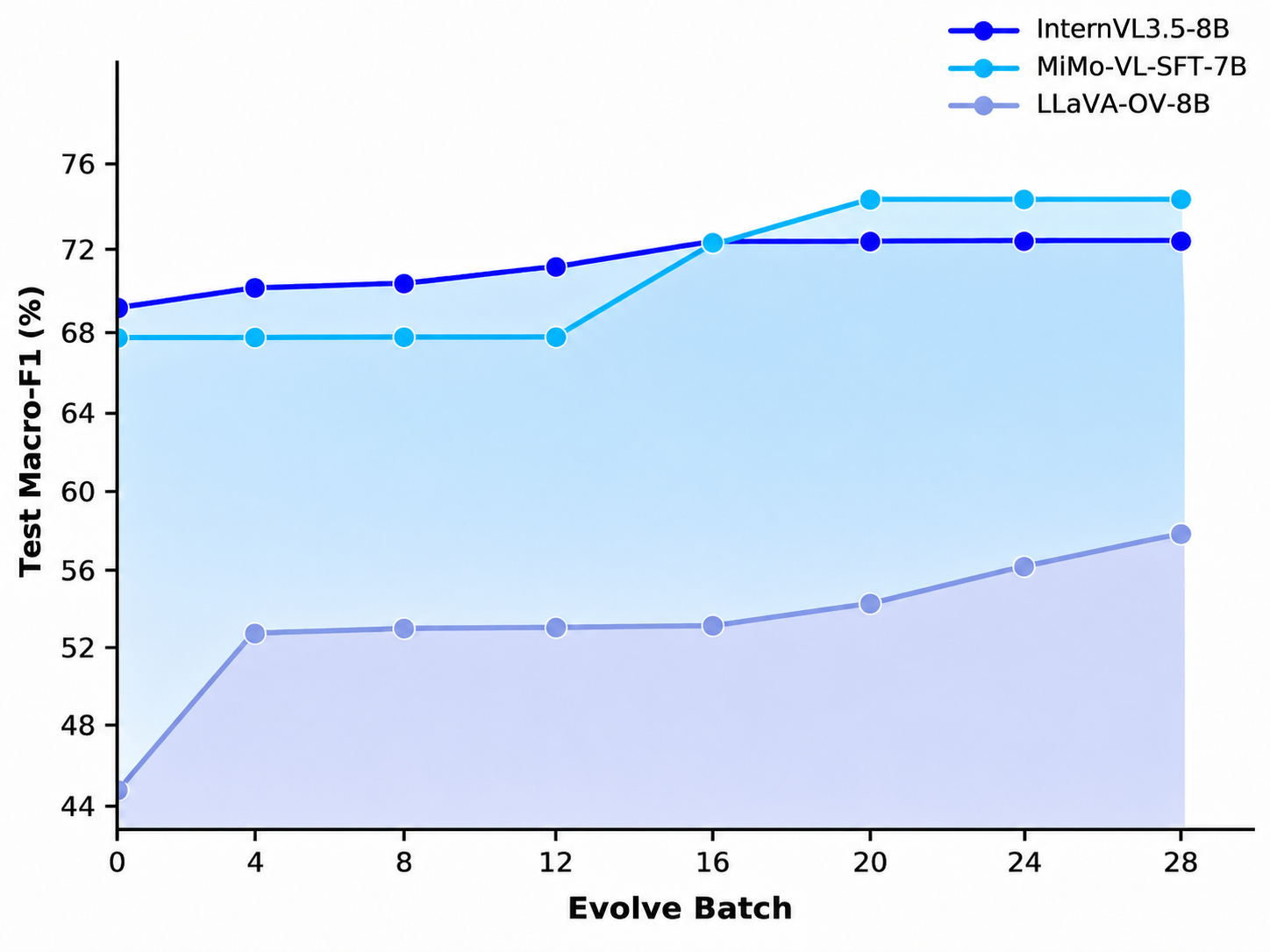}
        \par\small (b) Macro-F1
    \end{minipage}

    \caption{\lz{\textbf{Effect of evolution rounds.} We present the accuracy (left) and Macro-F1 (right) scores across three \textbf{backbones}: InternVL3.5-8B, MiMo-VL-SFT-7B, and LLaVA-OV-8B. All results are reported on the held-out test set.}}
    \label{fig:evolution_rounds}
\end{figure}

To investigate the impact of successive RSI updates on SPI performance, we evaluate three representative backbones, namely InternVL3.5-8B-Instruct, MiMo-VL-7B-SFT, and LLaVA-OneVision-2-8B-Instruct. Following the default configuration in Section~\ref{sec:implementation_details}, we assess ReMVP on the test set every four updates. Figure~\ref{fig:evolution_rounds}(a) and (b) present the corresponding Accuracy and Macro-F1 results, respectively.

All three backbones improve over their initial performance, although their evolution trajectories differ. These results suggest that beneficial policy revisions can emerge through accumulated feedback across multiple batches, with different backbones requiring different numbers of updates. Moreover, performance tends to plateau after a certain number of updates, indicating diminishing returns from continued evolution.

Overall, the consistent improvements in Accuracy and Macro-F1 demonstrate the effectiveness of recursively updating the skill generation policy for SPI tasks while keeping the backbone frozen.

\begin{figure*}[t]
    \centering
    \begin{minipage}[b]{0.63\textwidth}
        \centering
        \includegraphics[width=\linewidth]{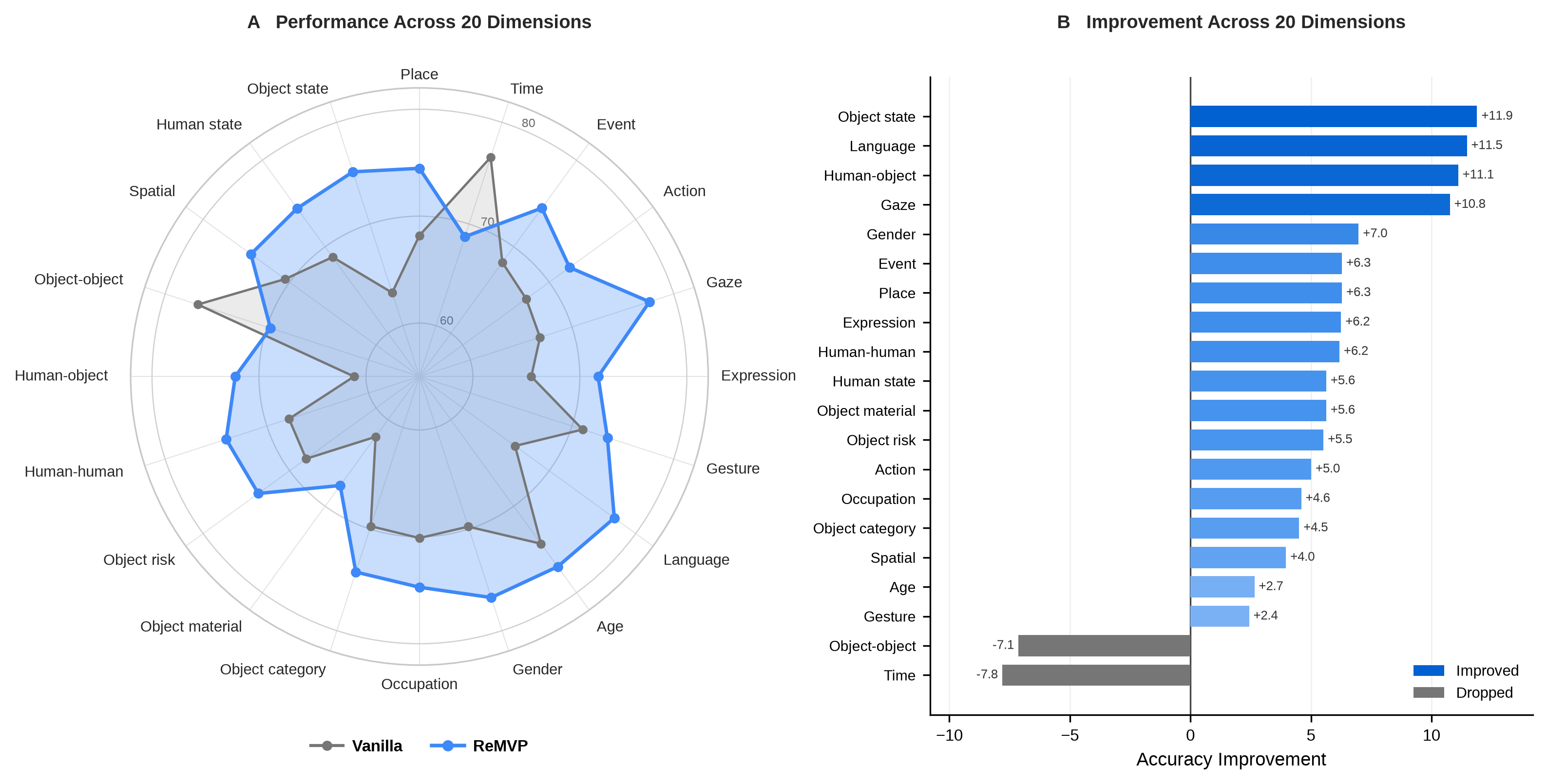}
        \par\small (a)
    \end{minipage}\hfill
    \begin{minipage}[b]{0.35\textwidth}
        \centering
        \includegraphics[width=\linewidth]{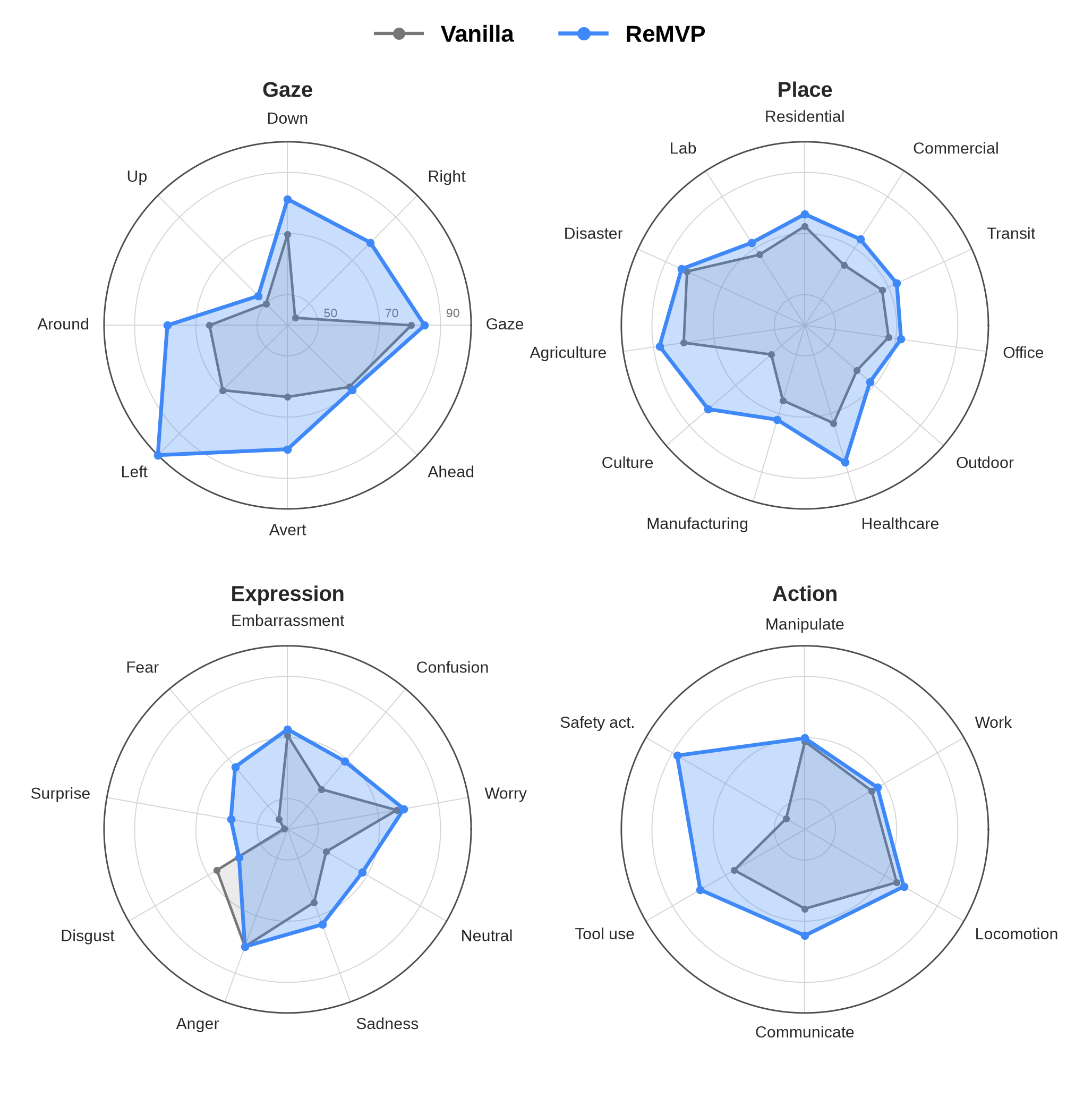}
        \par\small (b)
    \end{minipage}
    \caption{\textbf{Performance Analysis Across Video Attributes.} Results are evaluated on the attribute-annotated synthetic test subset. (a) Comparison of Vanilla and ReMVP across 20 video dimensions, showing action judgment accuracy and corresponding improvements in percentage points. (b) Detailed accuracy comparisons across Gaze, Place, Expression, and Action attributes, where each radar axis represents a specific attribute category. Blue and gray  respectively denote ReMVP and Vanilla.}
    \label{fig:l1_dimension_accuracy}
\end{figure*}

\subsection{Analysis of Video Dimensions}
\label{sec:exp_l1_analysis}

RobotEQ-Video~\cite{che2026roboteqvideo} organizes contextual factors relevant to SPI into a hierarchical world-state taxonomy comprising 20 dimensions and finer-grained Level-1 (L1) attributes. To assess the effectiveness of ReMVP across diverse video contexts, we analyze action judgment accuracy at both the dimension and attribute levels. We compare Vanilla and ReMVP using the same frozen Qwen3-VL-8B-Instruct backbone on the synthetic subset of the test set.

Figure~\ref{fig:l1_dimension_accuracy}(a) presents the results across the 20 dimensions. ReMVP outperforms Vanilla in 18 dimensions, with notable improvements in Object State, Language, Human-object Relations, and Gaze. Gains in Expression, Human-human Relations, and Human State further suggest improved reasoning in contexts involving social and interpersonal cues. However, performance declines in Object-object Relations and Time, indicating that the improvements are not uniform across all dimensions.
At the finer L1 level, Figure~\ref{fig:l1_dimension_accuracy}(b) shows improvements across multiple attributes of Gaze, Place, Expression, and Action, although certain attributes, such as Disgust, remain challenging. These findings suggest that ReMVP can leverage recursive policy refinement to improve action judgments across diverse contextual factors and fine-grained video attributes.

\newpage
\subsection{Validation for Beneficial Policy Evolution}
\label{sec:exp_val_gate}

\lz{
\begin{table}[b]
    \centering

    \caption{\lz{\textbf{Ablation analysis.} This table ablates the inner-batch validation and the policy update validation mechanisms to reveal their impact on ReMVP. Here, Avg denotes the mean of Accuracy and Macro-F1, and $\Delta$ indicates the Avg gain over Vanilla.}}
    \label{tab:val_gate}
    \fontsize{8}{10}\selectfont
    \setlength{\tabcolsep}{3pt}

    \begin{tabularx}{\columnwidth}{@{}>{\raggedright\arraybackslash}Xcccc@{\hspace{8pt}}}
        \specialrule{0.4pt}{0pt}{1pt}
        \specialrule{0.4pt}{0pt}{3pt}
        \textbf{Setting} & \textbf{Accuracy} & \textbf{Macro-F1} & \textbf{Avg} & $\boldsymbol{\Delta}$ \\
        \midrule[0.4pt]
        Vanilla & 70.28 & 69.31 & 69.80 & - \\
        w/o Inner Batch Validation & 73.65 & 71.72 & 72.69 & +2.89 \\
        w/o Update Validation & 74.81 & 72.55 & 73.68 & +3.88 \\
        \textbf{ReMVP} & \textbf{76.15} & \textbf{73.51} & \textbf{74.83} & \textbf{+5.03} \\
        \specialrule{0.4pt}{3pt}{1pt}
        \specialrule{0.4pt}{0pt}{0pt}
    \end{tabularx}
\end{table}
}

To guide policy evolution in a beneficial direction, ReMVP validates candidate policies at both the Recursive Improvement and Recursive Policy Update stages. During Recursive Improvement, Inner Batch Validation evaluates each policy revision on the current training batch, while during Recursive Policy Update, Update Validation evaluates the best policy from the current batch on a validation set before adopting it for subsequent updates. To assess their contributions, we conduct ablation studies by removing each validation stage separately and comparing the resulting variants with Vanilla and the complete ReMVP framework using Qwen3-VL-8B-Instruct on the RobotEQ-Video test set.

As shown in Table~\ref{tab:val_gate}, the complete ReMVP framework consistently outperforms both ablated variants across Accuracy, Macro-F1, and Avg. Removing either validation stage degrades performance, with a larger drop observed without Inner Batch Validation, highlighting the importance of both validation stages for effective policy refinement. Although both variants outperform Vanilla, the complete framework achieves the best overall performance, demonstrating the complementary contributions of the two validation stages.

\subsection{Cross-Model and Cross-Dataset Generalization}
\label{sec:exp_transfer}

To evaluate the cross-model transferability of evolved policies, we directly transfer policy $P_4$, evolved on Qwen3-VL-8B-Instruct, to InternVL3.5-8B-Instruct, Keye-VL-1.5-8B, and LLaVA-OneVision-2-8B-Instruct without additional evolution. We compare Transferred ReMVP with Vanilla and Own ReMVP, which independently evolves policies for each target model.

\begin{figure}[t]
    \centering
    \begin{minipage}[b]{0.49\columnwidth}
        \centering
        \includegraphics[height=0.5478\linewidth]{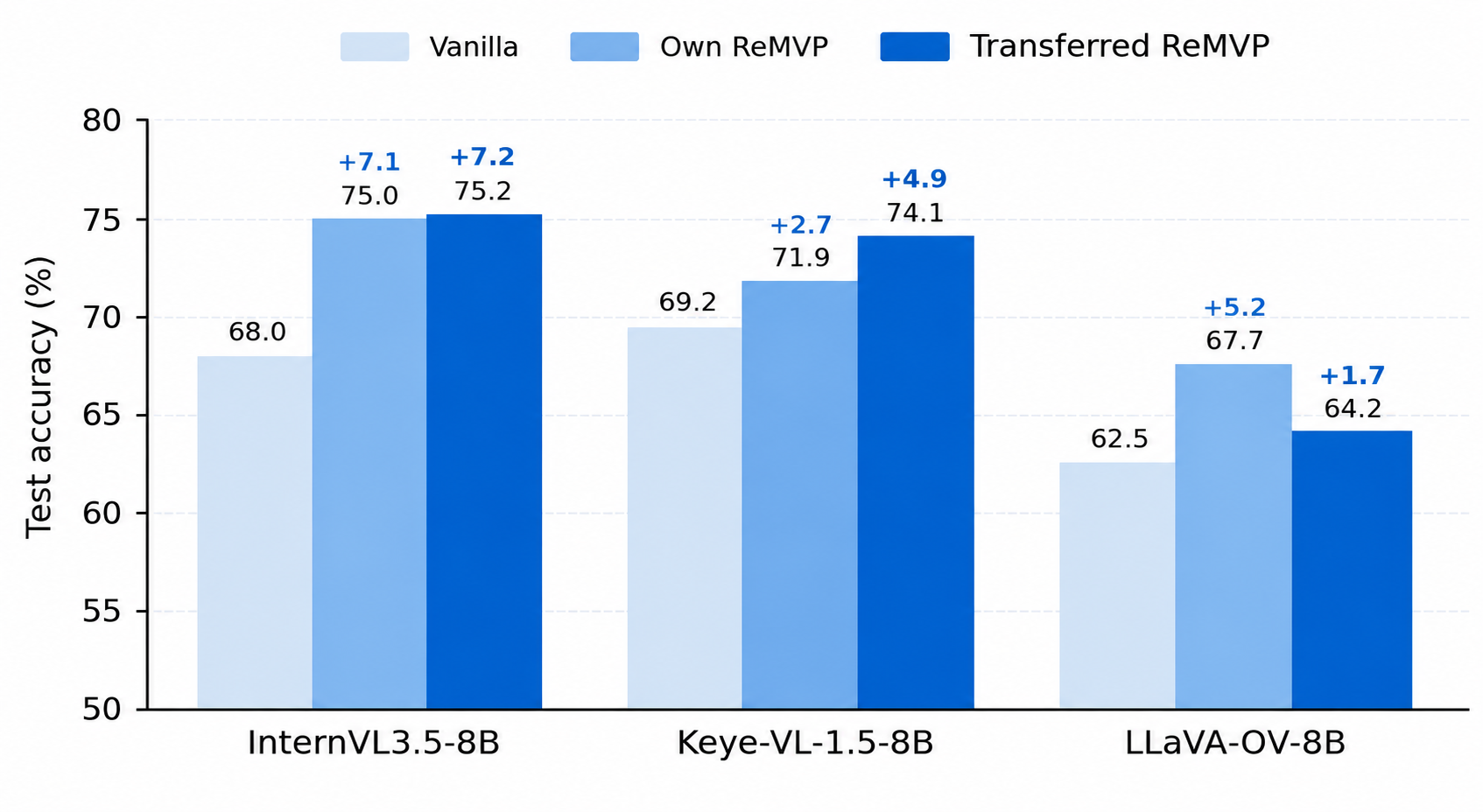}
        \par\small (a) Cross-model
    \end{minipage}\hfill
    \begin{minipage}[b]{0.49\columnwidth}
        \centering
        \includegraphics[height=0.576\linewidth]{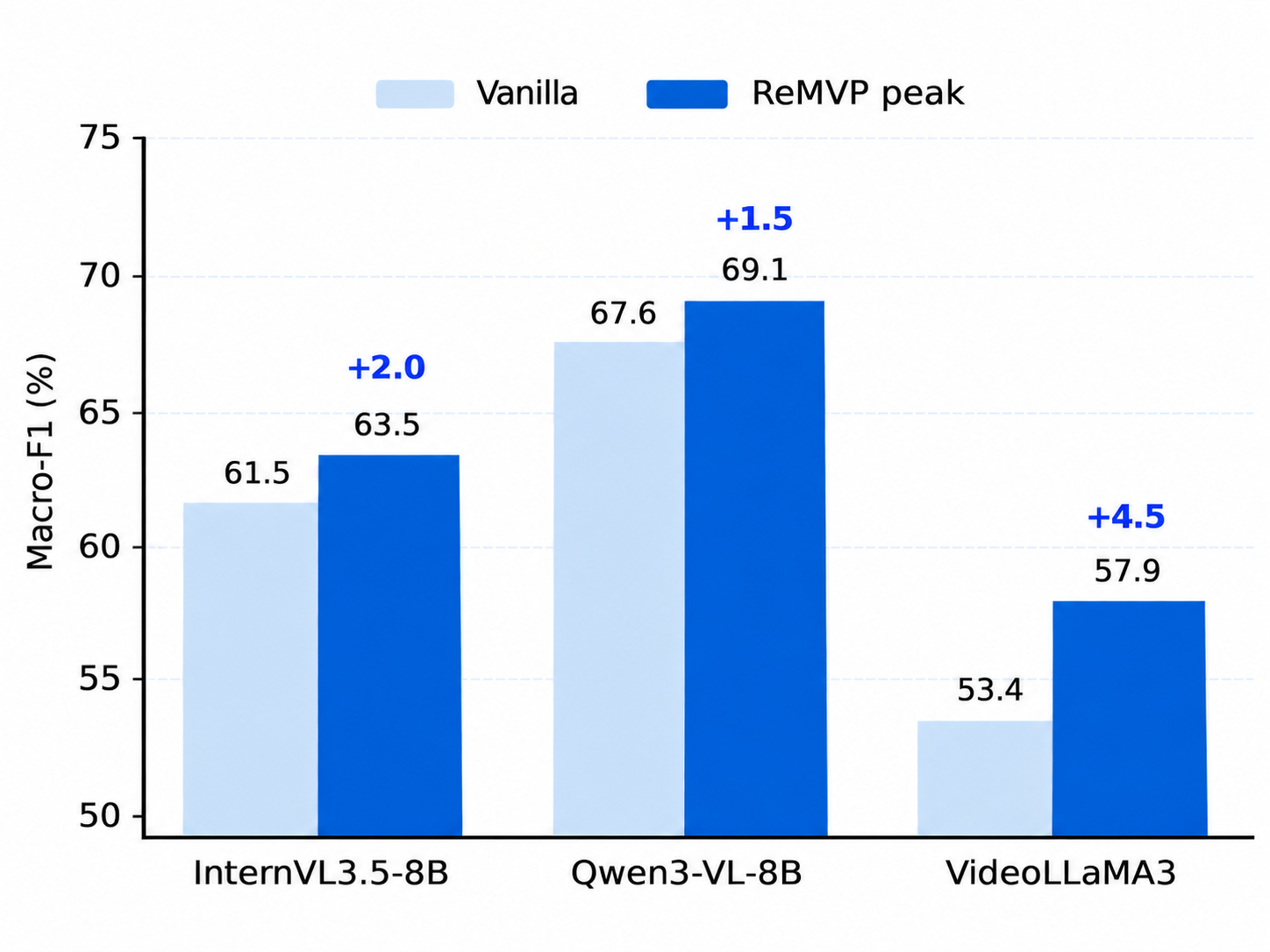}
        \par\small (b) Cross-dataset
    \end{minipage}

    \caption{{\textbf{Cross-model and cross-dataset generalization.}
(a) Cross-model transfer of evolved skill generation policies across different VLMs on RobotEQ-Video.
(b) Adaptation of ReMVP to the image-based RobotEQ 1.0 benchmark.}}
    \label{fig:cross_backbone_transfer}
\end{figure}

As shown in Fig.~\ref{fig:cross_backbone_transfer}(a), Transferred ReMVP consistently outperforms Vanilla across all three models. It also surpasses Own ReMVP on Keye and achieves slightly higher accuracy on InternVL, while remaining below Own ReMVP on LLaVA. These results highlight the transferability of evolved policies across different backbones.

We further evaluate ReMVP on RobotEQ 1.0~\cite{fang2026roboteq}, an image-based SPI benchmark with 978 training, 255 validation, and 300 test images. Following the RobotEQ-Video evolution settings, we refine policies using four images per batch and one revision per update, and evaluate performance using two-class Macro-F1, treating invalid predictions as false negatives for their ground-truth classes. As shown in Fig.~\ref{fig:cross_backbone_transfer}(b), ReMVP achieves higher peak Macro-F1 than Vanilla on InternVL3.5-8B, Qwen3-VL-8B, and VideoLLaMA3, with the largest gain on VideoLLaMA3.

Overall, these results demonstrate the potential of ReMVP to generalize across vision-language backbones and adapt to both video-based and image-based SPI tasks.

\subsection{Case Study}
\label{sec:exp_case_study}

\begin{figure}[b]
    \centering
    \includegraphics[width=\columnwidth]{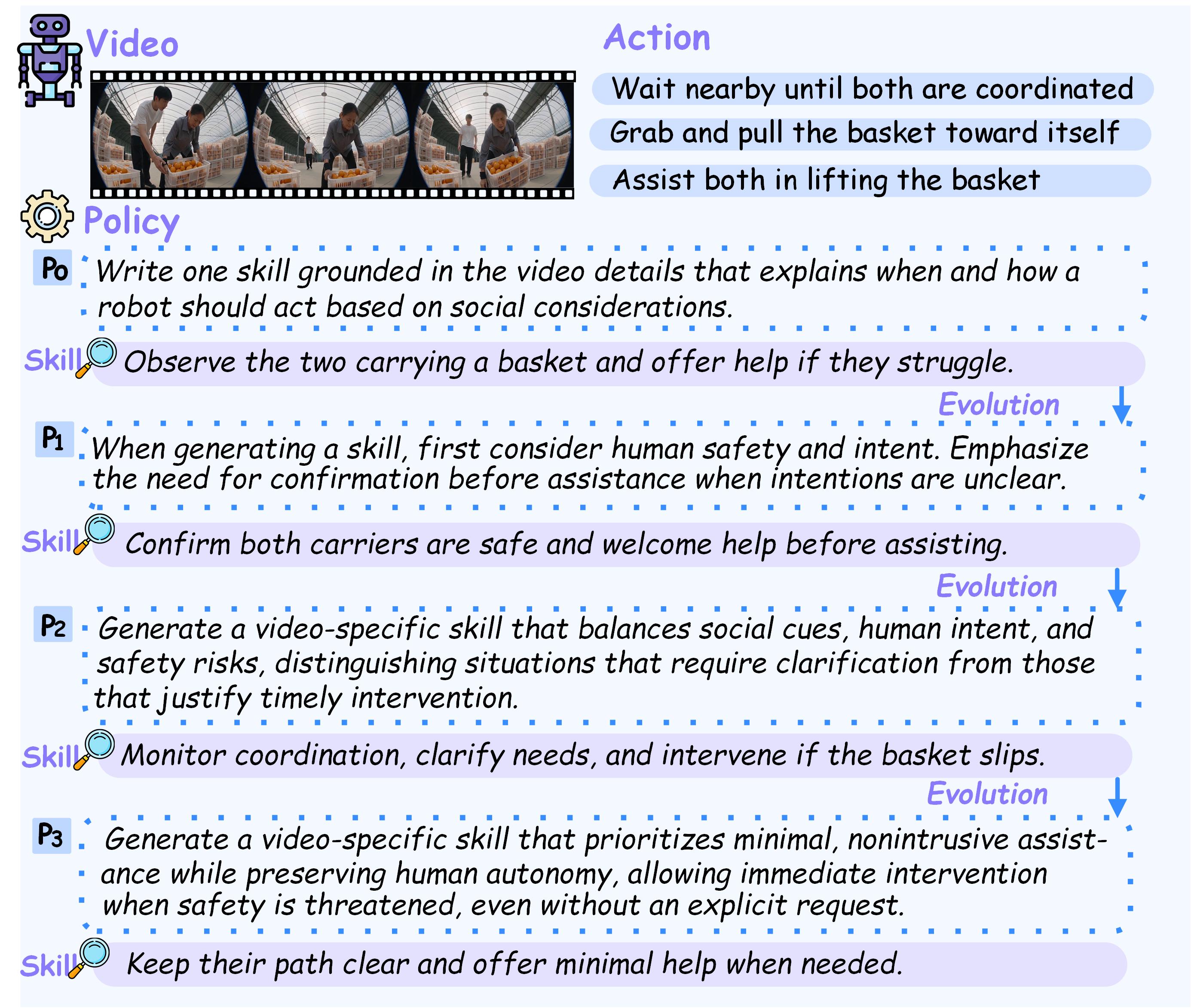}
    \caption{\textbf{Qualitative illustration of recursive policy and skill evolution.} Successive policy refinements and their corresponding skills progressively emphasize human intent, safety, and minimal assistance while preserving autonomy.}
    \label{fig:rsi_case_study}
    \label{fig:case_study}
\end{figure}

 To illustrate the evolution of skill generation policies and their corresponding skills, we present a representative case in Fig.~\ref{fig:case_study}. Starting from a general policy for video-grounded assistance, ReMVP progressively emphasizes confirmation of human intent before assistance , context-aware reasoning about safety and intervention timing, and minimal, timely intervention that respects human autonomy. This evolution reflects a shift from generic assistance toward a more nuanced understanding of social norms, particularly the importance of respecting human autonomy while responding to potential safety risks. Correspondingly, the generated skills become more specific to the observed interaction, providing increasingly refined guidance on when and how the robot should assist. This case illustrates how ReMVP can leverage recursive feedback to refine its skill generation policy and produce more socially appropriate skills.

\subsection{Comparison of Skill Utilization Strategies}
\label{sec:exp_fixed_skill}

To compare skill utilization strategies for SPI judgments, we evaluate ReMVP against a generate-then-retrieve baseline, termed Retrieved Skill. This baseline summarizes the RobotEQ training set into a fixed rule bank and retrieves relevant guidance for each test example. ReMVP generates video-specific skills using a recursively evolved policy. All three settings, including Vanilla, use the same frozen Qwen3-VL-8B-Instruct backbone and evaluation configuration.

\begin{table}[h]
    \centering

    \caption{\lz{\textbf{Comparison of skill utilization paradigms.} Vanilla performs action judgments without skills, Retrieved Skill uses retrieved experiences as guidance, and ReMVP generates video-specific skills through recursive policy improvement.}}
    \label{tab:fixed_skill_comparison}
    \fontsize{8.5}{10.5}\selectfont
    \setlength{\tabcolsep}{3pt}
    \renewcommand{\arraystretch}{1.12}
    \begin{tabularx}{\columnwidth}{@{}>{\raggedright\arraybackslash}Xcccc@{}}
        \specialrule{0.4pt}{0pt}{1pt}
        \specialrule{0.4pt}{0pt}{3pt}
        \textbf{Setting} & \textbf{Accuracy} & \textbf{Macro-F1} & \textbf{Avg} & $\boldsymbol{\Delta}$ \\
        \midrule[0.4pt]
        Vanilla & 70.28 & 69.31 & 69.80 & - \\
        Retrieved Skill & 72.08 & 67.35 & 69.72 & -0.08 \\
        \addlinespace[2pt]
        \textbf{ReMVP} & \textbf{76.15} & \textbf{73.51} & \textbf{74.83} & \textbf{+5.03} \\
        \specialrule{0.4pt}{3pt}{1pt}
        \specialrule{0.4pt}{0pt}{0pt}
    \end{tabularx}
\end{table}
Table~\ref{tab:fixed_skill_comparison} shows that Retrieved Skill improves Accuracy over Vanilla but lowers Macro-F1, leaving Avg slightly below the baseline. Retrieved guidance therefore does not improve performance consistently across the two action classes. ReMVP achieves the highest Accuracy, Macro-F1, and Avg among the three settings. These results favor video-specific guidance generated under an evolved policy over retrieval from this fixed rule bank.

\raggedbottom
\section{Conclusion}
\label{sec:conclusion}

We propose ReMVP, a training-free framework that recursively refines skill generation policies using feedback from action judgments. By leveraging video-specific generated skills, ReMVP enhances models' ability to reason about complex and diverse video contexts. This method substantially improves embodied agents' social behavior reasoning in complex dynamic videos. It is the first to propose a skill-cue prompting enhancement framework for SPI and the first to apply the RSI mechanism to this domain. Extensive experiments on RobotEQ-Video demonstrate its effectiveness and further validate its generalization capabilities across different models and datasets.

\section{Limitation}
\label{sec:limitations}

ReMVP requires labeled judgments and additional computation for policy evolution. Our evaluation is limited to recorded videos and does not assess real-world robot execution. Performance varies across contextual dimensions, and retrospectively selected test peaks may overestimate generalization. Future work will address these limitations.
\newpage
\section*{LLM Usage Statement}

\lz{

ChatGPT and Codex were used to support methodological brainstorming, manuscript drafting and editing, literature search and citation formatting, and figure and table preparation. These tools were not involved in any inference or evaluation procedures reported in this paper.

}

{
    \small
    \bibliographystyle{ieeenat_fullname}
    \bibliography{main}
}

\clearpage
\appendix
\raggedbottom

\section{Methodological Details}
\label{app:prompts}
This section provides additional methodological details.

\subsection{Initial Skill-Generation Policy}
\label{app:initial_prompt}
\begin{promptbox}{Initial Skill-Generation Policy}
Write one skill grounded in the video details that explains when and how a robot should act based on social considerations.
\end{promptbox}

\subsection{Skill Output Format}
\label{app:skill_format}
The skill writer returns one video-level skill in the following JSON format.
\begin{promptbox}{Skill Output}
\begin{lstlisting}[style=skilljson]
{
  "skill": "Observe the two carrying a basket and offer help if they struggle."
}
\end{lstlisting}
\end{promptbox}

\subsection{Rollout Information}
\label{app:rollout_feedback}
The rollout information provides the proposal agent with feedback on the current prompt's performance within each training window. Specifically, it includes the following information:

\begin{promptbox}{Rollout Information}
\begin{itemize}[leftmargin=*,itemsep=5pt,topsep=2pt,parsep=0pt]
    \item \textbf{Video:} The input video.
    \item \textbf{Extracted Skill:} The episode-level skill generated from the video using the current prompt.
    \item \textbf{Candidate Actions:} All candidate actions associated with the video.
    \item \textbf{Ground-Truth Labels:} The ground-truth label for each candidate action.
    \item \textbf{No-Skill Predictions:} Action judgments produced without the extracted skill.
    \item \textbf{Skill-Conditioned Predictions:} Action judgments produced with the extracted skill, together with the corresponding changes in prediction correctness.
    \item \textbf{Gain Score:} The net improvement in action judgments, computed as:
\end{itemize}
\[
\begin{aligned}
\mathrm{Recover}_n &= \sum_{i\in D_n}\sum_{j=1}^{m_i} (1-c_{ij})c_{ij}^{s},\\[6pt]
\mathrm{Failure}_n &= \sum_{i\in D_n}\sum_{j=1}^{m_i} c_{ij}(1-c_{ij}^{s}),\\[6pt]
\mathrm{Gain}_n &= \mathrm{Recover}_n-\mathrm{Failure}_n.
\end{aligned}
\]
Here, $\mathrm{Recover}_n$ counts the incorrect judgments corrected by the extracted skill, while $\mathrm{Failure}_n$ counts the previously correct judgments that become incorrect after incorporating the skill.
\end{promptbox}

\subsection{Policy Evolution Proposal Prompt}
\label{app:proposal_prompt}
\begin{promptbox}{Policy Evolution Proposal Prompt}
You are a prompt optimization agent for Social Proactive Intelligence (SPI) tasks.

Evolve the current skill extraction prompt based on evaluation feedback to improve the quality of generated skills.

\textbf{Requirements}
\begin{enumerate}[leftmargin=*,itemsep=3pt,topsep=2pt,parsep=0pt]
\item Identify the most critical weakness in the current prompt.
\item Introduce new strategies or instructions when they address the identified weakness.
\item Replace, refine, or remove less effective instructions to accommodate improvements.
\item Preserve effective guidance for social understanding, proactive decisions, and generalization.
\item Keep the revised prompt within 100 words.
\item Prioritize meaningful improvements over superficial rewording.
\item Do not add refusal, blank-skill, or anti-observation rules. The Revised Prompt must never tell the skill writer to output ``(no episode\_skill)'', skip writing a skill, return empty text, or forbid assessment/notification/observation.
\end{enumerate}

\begin{minipage}{\linewidth}
\textbf{Input}
\begin{itemize}[leftmargin=*,itemsep=2pt,topsep=2pt,parsep=0pt]
\item Current Prompt: \texttt{\{CURRENT\_PROMPT\}}
\item Extracted Skills: \texttt{\{SKILLS\}}
\item Evaluation Feedback: \texttt{\{FEEDBACK\}}
\end{itemize}
\end{minipage}

\begin{minipage}{\linewidth}
\textbf{Output}
\begin{itemize}[leftmargin=*,itemsep=2pt,topsep=2pt,parsep=0pt]
\item \textbf{Analysis:} The primary weakness.
\item \textbf{Proposal:} The proposed improvement and its rationale.
\item \textbf{Revised Prompt:} A complete, self-contained prompt of no more than 100 words.
\end{itemize}
\end{minipage}

These instructions are for you (the proposal agent) only. Do not copy them into the Revised Prompt.
\end{promptbox}

\section{Implementation Details}
\label{app:implementation}
\subsection{Video Frames}
\label{app:video_frames}
We adopt uniform frame sampling for video inputs. Specifically, the skill extraction and action evaluation stages use 8 frames per video, while the prompt optimization stage uses up to 4 videos from each training window, with 4 frames sampled per video.

\subsection{Dataset Splits}
\label{app:dataset_splits}
All experiments are conducted on the Robot-EQ dataset. We randomly select 100 videos from the original validation set for candidate selection and merge the remaining 332 videos into the training set, resulting in 1,833 training videos with 15,004 annotated action options. The test set consists of 213 videos with 1,723 action options, including 132 synthetic videos (1,170 action options) and 81 real-world videos (553 action options). Table~\ref{tab:supp_dataset_splits} summarizes the dataset statistics.

\begin{table}[!ht]
    \centering
    \caption{\textbf{Dataset split statistics.} Video and annotated action-option counts for training, validation, and testing.}
    \label{tab:supp_dataset_splits}
    \small
    \setlength{\tabcolsep}{8pt}
    \renewcommand{\arraystretch}{1.15}
    \begin{tabular*}{\columnwidth}{@{\extracolsep{\fill}}lcc@{}}
        \specialrule{0.4pt}{0pt}{1pt}
        \specialrule{0.4pt}{0pt}{3pt}
        \textbf{Split} & \textbf{Videos} & \textbf{Action Options} \\
        \midrule[0.4pt]
        Training & 1,833 & 15,004 \\
        Validation & 100 & 817 \\
        Test & 213 & 1,723 \\
        \specialrule{0.4pt}{3pt}{1pt}
        \specialrule{0.4pt}{0pt}{0pt}
    \end{tabular*}
\end{table}

Each action option is annotated with a binary label, either \emph{Should} or \emph{Should Not}.

\subsection{Hardware and Software}
\label{app:hardware_software}
All experiments are conducted on NVIDIA H800 GPUs (80 GB), with each 7B/8B model running on a single GPU. Our implementation is based on Python 3.12, PyTorch, Hugging Face Transformers, and \texttt{qwen-vl-utils}. We configure model-specific GPU memory limits, as summarized in Table~\ref{tab:supp_gpu_memory}. Video decoding is performed using torchvision.

\begin{table}[!ht]
    \centering
    \caption{\textbf{GPU memory limits for different models.}}
    \label{tab:supp_gpu_memory}
    \fontsize{8}{10}\selectfont
    \setlength{\tabcolsep}{3pt}
    \renewcommand{\arraystretch}{1.12}
    \begin{tabular*}{\columnwidth}{@{\extracolsep{\fill}}lc@{}}
        \specialrule{0.4pt}{0pt}{1pt}
        \specialrule{0.4pt}{0pt}{3pt}
        \textbf{Model} & \textbf{GPU Memory Cap} \\
        \midrule[0.4pt]
        InternVL3.5-4B-Instruct & 14 GiB \\
        MiMo-VL-7B-RL & 18 GiB \\
        MiMo-VL-7B-SFT & 18 GiB \\
        VideoLLaMA3-7B & 24--40 GiB \\
        Qwen3-VL-8B-Instruct & 16 GiB \\
        Keye-VL-1.5-8B & 20 GiB \\
        InternVL3.5-8B-Instruct & 20 GiB \\
        LLaVA-OneVision-2-8B-Instruct & 20 GiB \\
        \specialrule{0.4pt}{3pt}{1pt}
        \specialrule{0.4pt}{0pt}{0pt}
    \end{tabular*}
\end{table}

\Needspace{240pt}
\subsection{Default Hyperparameters}
\label{app:default_hyperparameters}
\begin{table}[H]
    \centering
    \caption{\textbf{Default hyperparameters.}}
    \label{tab:supp_default_hyperparameters}
    \fontsize{8}{10}\selectfont
    \setlength{\tabcolsep}{3pt}
    \renewcommand{\arraystretch}{1.15}
    \begin{tabular*}{\columnwidth}{@{\extracolsep{\fill}}lc@{}}
        \specialrule{0.4pt}{0pt}{1pt}
        \specialrule{0.4pt}{0pt}{3pt}
        \textbf{Item} & \textbf{Value} \\
        \midrule[0.4pt]
        Model & Qwen3-VL-8B-Instruct \\
        GPU & $1\times$ H800 80 GB \\
        HF memory cap & 16 GiB, bf16 \\
        Extra rewrites & $extra=1$ \\
        Batch / window & 4 videos \\
        Frames (skill / judge) & 8 uniform \\
        Frames (proposal) & 4 \\
        Proposal videos & 4 \\
        Proposal temperature & 1.0 \\
        Skill \texttt{max\_new\_tokens} & 512 \\
        Judge \texttt{max\_new\_tokens} & 384 \\
        Proposal \texttt{max\_new\_tokens} & 1024 \\
        Inference batch size & 1 \\
        Early stopping patience & 10 validation checks \\
        \specialrule{0.4pt}{3pt}{1pt}
        \specialrule{0.4pt}{0pt}{0pt}
    \end{tabular*}
\end{table}

Both the main results in Sec.~5.1 and the analysis of RSI evolution rounds in Sec.~5.2 follow the default experimental settings summarized in Table~\ref{tab:supp_default_hyperparameters}. For the latter, each point in the curve represents the test-set performance of the evolved policy retained at the corresponding evolution step. Specifically, training windows provide the signal for policy evolution, while the validation set determines whether an updated policy is retained. The test set is used solely for evaluation and does not participate in policy evolution or selection.

\Needspace{6\baselineskip}
\subsection{RobotEQ-video Dataset and Generation Dimensions}
\label{app:dataset_dimensions}

\paragraph{RobotEQ-video.}
RobotEQ-video is a video-based benchmark designed to evaluate whether robot actions are socially appropriate in a given context. Each sample consists of a short video and a set of candidate actions, where each action is annotated with a binary label: \emph{Should} or \emph{Should Not}. The benchmark requires models to reason about the visual context rather than relying solely on action descriptions, as seemingly helpful or polite actions may be inappropriate due to social boundaries, ownership constraints, timing, or lack of consent.

The dataset comprises two sources: \textbf{synthetic videos} and \textbf{real-world videos}. The synthetic videos are generated based on a taxonomy of 20 dimensions, covering various aspects of social interactions, environmental contexts, human behaviors, and object relationships. Each synthetic video is associated with dimension-level annotations and corresponding L1 category labels. The synthetic video collection contains 1,218 videos. The real-world collection consists of 928 videos captured in natural environments, including egocentric perspectives.

\paragraph{Generation Dimensions.}
Synthetic videos are constructed based on 20 generation dimensions, as summarized in Table~\ref{tab:supp_generation_dimensions}. Each synthetic video is associated with dimension-level labels.

\begin{table}[H]
    \centering
    \caption{\textbf{The 20 generation dimensions used in RobotEQ-video.}}
    \label{tab:supp_generation_dimensions}
    \fontsize{8}{10}\selectfont
    \setlength{\tabcolsep}{6pt}
    \renewcommand{\arraystretch}{1.08}
    \begin{tabular*}{\columnwidth}{@{\extracolsep{\fill}}cl@{}}
        \specialrule{0.4pt}{0pt}{1pt}
        \specialrule{0.4pt}{0pt}{3pt}
        \textbf{No.} & \textbf{Generation Dimension} \\
        \midrule[0.4pt]
        1 & Place \\
        2 & Time \\
        3 & Event \\
        4 & Action \\
        5 & Gaze \\
        6 & Expression \\
        7 & Gesture \\
        8 & Linguistic Content \\
        9 & Age \\
        10 & Gender \\
        11 & Occupation \\
        12 & Object Category \\
        13 & Object Material \\
        14 & Object Risk \\
        15 & Human--Human Relation \\
        16 & Human--Object Relation \\
        17 & Object--Object Relation \\
        18 & Spatial Relation \\
        19 & Human State \\
        20 & Object State \\
        \specialrule{0.4pt}{3pt}{1pt}
        \specialrule{0.4pt}{0pt}{0pt}
    \end{tabular*}
\end{table}

\Needspace{6\baselineskip}
\subsection{Ablation Study on Validation Strategies}
\label{app:validation_ablations}
We investigate the effectiveness of the two validation strategies in ReMVP: inner-batch validation and update validation. Inner-batch validation selects the candidate policy with the highest score within each training window, while update validation determines whether to retain the candidate based on its performance on the validation set. Table~\ref{tab:supp_validation_ablations} summarizes the configurations of the different variants.

\begin{table}[H]
    \centering
    \caption{\textbf{Configurations of the validation ablations.}}
    \label{tab:supp_validation_ablations}
    \fontsize{8}{10}\selectfont
    \setlength{\tabcolsep}{3pt}
    \renewcommand{\arraystretch}{1.18}
    \begin{tabular}{@{}>{\raggedright\arraybackslash}p{0.28\columnwidth}>{\raggedright\arraybackslash}p{0.29\columnwidth}>{\raggedright\arraybackslash}p{\dimexpr0.43\columnwidth-12pt\relax}@{}}
        \specialrule{0.4pt}{0pt}{1pt}
        \specialrule{0.4pt}{0pt}{3pt}
        \textbf{Method} & \textbf{Inner-Batch Validation} & \textbf{Update Validation} \\
        \midrule[0.4pt]
        ReMVP (Full) & Best candidate within each window & Retain the historically best policy on the validation set \\
        \addlinespace[2pt]
        w/o Inner-Batch Validation & Last rewritten policy & Retain the historically best policy on the validation set \\
        \addlinespace[2pt]
        w/o Update Validation & Best candidate within each window & Disabled \\
        \addlinespace[2pt]
        w/o Both & Last rewritten policy & Disabled \\
        \specialrule{0.4pt}{3pt}{1pt}
        \specialrule{0.4pt}{0pt}{0pt}
    \end{tabular}
\end{table}

\Needspace{6\baselineskip}
\subsection{Detail of Generalization Experiment}
\label{app:generalization_details}
\subsubsection{Detail of Cross-Model Generalization}
\label{app:cross_model_generalization}

We investigate the cross-model transferability of skill-extraction prompts on RobotEQ-video. We first apply ReMVP to Qwen3-VL-8B-Instruct using the default settings and select the prompt achieving the best validation performance, denoted as $P_4$. We then freeze $P_4$ and directly transfer it to three target VLMs: InternVL3.5-8B-Instruct, Keye-VL-1.5-8B, and LLaVA-OneVision-2-8B-Instruct. Each target model uses $P_4$ to extract episode-specific skills and performs skill-conditioned action judgments. All models are evaluated on the same RobotEQ-video test split without further prompt evolution or model weight updates.

\begin{promptbox}{Selected Prompt ($P_4$)}
Generate an episode-specific skill that establishes a context-aware decision rule and explicit action ordering. First assess social norms, implied consent, relational dynamics, and micro-social cues, then provide minimal, nonintrusive assistance only when socially appropriate or safety is endangered, even without an explicit request.
\end{promptbox}

\subsubsection{Detail of Cross-Dataset Generalization}
\label{app:cross_dataset_generalization}

We further evaluate the cross-dataset transferability of ReMVP on \textbf{Robot-Image (Robot-EQ 1.0)}, a first-person benchmark for assessing the appropriateness of robot actions in social contexts. Each sample consists of an egocentric image, a role or question description, and one or more candidate actions. Each action is labeled as either \emph{proper} or \emph{improper}, which we map to the binary labels \emph{Should} and \emph{Should Not}, respectively, consistent with RobotEQ-video. The model is required to assess the appropriateness of each candidate action based on the visual context.

We adopt the official dataset split, consisting of three mutually exclusive subsets with a total of 1,533 images and 2,032 annotated action options. The dataset statistics are summarized in Table~\ref{tab:supp_roboteq_image_splits}.

\begin{table}[H]
    \centering
    \caption{\textbf{Dataset statistics of Robot-EQ 1.0.}}
    \label{tab:supp_roboteq_image_splits}
    \fontsize{8}{10}\selectfont
    \setlength{\tabcolsep}{3pt}
    \renewcommand{\arraystretch}{1.15}
    \begin{tabular*}{\columnwidth}{@{\extracolsep{\fill}}lcccc@{}}
        \specialrule{0.4pt}{0pt}{1pt}
        \specialrule{0.4pt}{0pt}{3pt}
        \textbf{Split} & \textbf{Images} & \shortstack{\textbf{Action}\\\textbf{Options}} & \textbf{Proper} & \textbf{Improper} \\
        \midrule[0.4pt]
        Training & 978 & 1,280 & 859 & 421 \\
        Validation & 255 & 362 & 225 & 137 \\
        Test & 300 & 390 & 256 & 134 \\
        \specialrule{0.4pt}{3pt}{1pt}
        \specialrule{0.4pt}{0pt}{0pt}
    \end{tabular*}
\end{table}

Given the class imbalance in Robot-EQ 1.0, where 256 out of 390 test action options (65.6\%) are labeled as \emph{proper}, we adopt two-class macro-F1 as the primary evaluation metric to account for performance on both classes.

For cross-dataset transfer, we follow the ReMVP configuration used for Qwen3-VL-8B on RobotEQ-video with $extra=1$, adapting the framework to image-based inputs. We retain the same initial policy $P_0$ and inner-batch candidate selection strategy, while disabling validation-based keep-best selection. Table~\ref{tab:supp_cross_dataset_settings} summarizes the experimental configurations for the two datasets.

\begin{table}[H]
    \centering
    \caption{\textbf{Comparison of experimental settings between RobotEQ-video and Robot-Image (Robot-EQ 1.0).}}
    \label{tab:supp_cross_dataset_settings}
    \fontsize{8}{10}\selectfont
    \setlength{\tabcolsep}{3pt}
    \renewcommand{\arraystretch}{1.18}
    \begin{tabular}{@{}>{\raggedright\arraybackslash}p{0.25\columnwidth}>{\raggedright\arraybackslash}p{\dimexpr0.375\columnwidth-6pt\relax}>{\raggedright\arraybackslash}p{\dimexpr0.375\columnwidth-6pt\relax}@{}}
        \specialrule{0.4pt}{0pt}{1pt}
        \specialrule{0.4pt}{0pt}{3pt}
        \textbf{Setting} & \textbf{RobotEQ-video (Ours)} & \textbf{Robot-Image (Robot-EQ 1.0)} \\
        \midrule[0.4pt]
        Input & Video (8 frames) & Image (1 frame) \\
        \addlinespace[2pt]
        Window size & 4 videos & 4 images \\
        \addlinespace[2pt]
        Extra rewrites & $extra=1$ & $extra=1$ \\
        \addlinespace[2pt]
        Evolution windows & 459 (1,833 videos) & 245 (978 images) \\
        \specialrule{0.4pt}{3pt}{1pt}
        \specialrule{0.4pt}{0pt}{0pt}
    \end{tabular}
\end{table}

\Needspace{6\baselineskip}
\subsection{Details of Experience Generation and Retrieval}
\label{app:experience_generation_retrieval}

To further investigate the generalization capability of ReMVP, we conduct an additional experiment using a retrieval-based experience bank. We construct the experience bank from the training set by providing an \textbf{expert model} with training samples, their corresponding model response trajectories, and ground-truth annotations. The expert model distills these examples into reusable experiences that capture generalizable social norms and behavioral guidelines.

For experience retrieval, we employ \textbf{Qwen3-VL-Embedding} to encode videos and experience texts into a shared embedding space. We compute the cosine similarity between each query video and the experiences in the bank to identify the most relevant experiences:
\begin{equation}
    s(v,e_i)=\frac{\mathbf{z}_v^{\top}\mathbf{z}_{e_i}}
    {\lVert\mathbf{z}_v\rVert_2\,\lVert\mathbf{z}_{e_i}\rVert_2}.
    \label{eq:supp_experience_similarity}
\end{equation}
Here, $\mathbf{z}_v$ denotes the video embedding and $\mathbf{z}_{e_i}$ denotes the embedding of the $i$-th experience. The retrieved experiences are incorporated into the action evaluation process to provide additional guidance for decision-making. The experience bank is constructed exclusively from the training set, without using test-set annotations.

\Needspace{9\baselineskip}
Four representative examples of the generated experiences are presented below:
\begin{promptbox}{Representative Experiences}
\begin{enumerate}[leftmargin=*,itemsep=5pt,topsep=4pt,parsep=0pt]
    \item Respect Others' Feelings and Personal Boundaries.
    \item Listen Attentively and Respond at Appropriate Moments.
    \item Preserve Others' Dignity in Social Interactions.
    \item Express Politeness According to Social Context.
\end{enumerate}
\end{promptbox}

\end{document}